\documentclass[conference]{IEEEtran}
\IEEEoverridecommandlockouts
\usepackage{cite}
\usepackage{amsmath,amssymb,amsfonts}
\usepackage{algorithmic}
\usepackage{graphicx}
\usepackage{textcomp}
\usepackage{xcolor}
\usepackage{booktabs}
\usepackage{subcaption}
\def\BibTeX{{\rm B\kern-.05em{\sc i\kern-.025em b}\kern-.08em
    T\kern-.1667em\lower.7ex\hbox{E}\kern-.125emX}}
\begin{document}

\title{Synthetic Time Series Data Generation for Healthcare Applications: A PCG Case Study\\

}
\author{\IEEEauthorblockN{Ainaz Jamshidi}
\IEEEauthorblockA{\textit{Department of Information Systems} \\
\textit{University of Maryland Baltimore County}\\
Baltimore, USA \\
ainazj1@umbc.edu}
\and
\IEEEauthorblockN{Muhammad Arif}
\IEEEauthorblockA{\textit{Department of Information Systems} \\
\textit{Colorado State University}\\
Pueblo, USA \\
muhammad.arif@csupueblo.edu}
\and
\IEEEauthorblockN{Sabir Ali Kalhoro}
\IEEEauthorblockA{\textit{Department of Electrical Engineering} \\
\textit{University of Colorado, Colorado Springs}\\
Colorado, USA \\
skalhoro@uccs.edu}
\and
\IEEEauthorblockN{Alexander Gelbukh}
\IEEEauthorblockA{\textit{Department of Computing Research Center} \\
\textit{Instituto Politécnico Nacional}\\
New Mexico, USA \\
gelbukh@cic.ipn.mx}
}

\maketitle

\begin{abstract}
The generation of high-quality medical time series data is essential for advancing healthcare diagnostics and safeguarding patient privacy. Specifically, synthesizing realistic phonocardiogram (PCG) signals offers significant potential as a cost-effective and efficient tool for cardiac disease pre-screening.
Despite its potential, the synthesis of PCG signals for this specific application received limited attention in research.
In this study, we employ and compare three state-of-the-art generative models from different categories — WaveNet, DoppelGANger, and DiffWave — to generate high-quality PCG data. We use data from the George B. Moody PhysioNet Challenge 2022.
Our methods are evaluated using various metrics widely used in the previous literature in the domain of time series data generation, such as mean absolute error and maximum mean discrepancy. Our results demonstrate that the generated PCG data closely resembles the original datasets, indicating the effectiveness of our generative models in producing realistic synthetic PCG data.
In our future work, we plan to incorporate this method into a data augmentation pipeline to synthesize abnormal PCG signals with heart murmurs, in order to address the current scarcity of abnormal data. We hope to improve the robustness and accuracy of diagnostic tools in cardiology, enhancing their effectiveness in detecting heart murmurs.

\end{abstract}


\section{Introduction}
\textbf{Significance and Motivation.
}Synthesizing data through generative models refers to the application of advanced artificial intelligence models to create high-quality artificial data. This field of study involves creating artificial data that mimic real patient information, medical records, or diagnostic images. These synthetic data serve various purposes, such as training machine learning models, conducting research, and addressing data privacy concerns.
By generating realistic and high-quality synthetic healthcare data, researchers and practitioners can overcome limitations related to data availability, privacy regulations, and data diversity. Ultimately, the successful generation of realistic medical time series synthetic data will contribute significantly to the advancement of healthcare diagnostic tools and eventually the enhancement of patient care.

\textbf{Problem Statement and the Methodology.
}To the best of our knowledge, no existing studies focus on the generation of high-quality PCG signals using generative models. In this study, we analyze the performance of state-of-the-art generative models across three distinct categories: autoregressive models, generative adversarial networks (GANs), and diffusion-based approaches. Predictive modeling, and distribution analysis are used to assess the quality of generating (QoG) of the synthesized data.

\section{Related Work}
Generating synthetic medical time series data is a crucial and challenging area of research due to its sequential nature. Esteban et al. proposed \cite{esteban2017real} two types of generative models called Recurrent GAN (RGAN) and Recurrent Conditional GAN (RCGAN) for generating realistic real-valued multi-dimensional time series data, focusing on their application in the medical domain. RGANs and RCGANs utilize recurrent neural networks (RNNs) in both the generator and the discriminator. In the case of RCGANs, these RNNs are conditioned on auxiliary information.
Hazra et al. introduced \cite{hazra2020synsiggan} SynSigGAN, a GAN model designed to generate synthetic biomedical signals. The GAN incorporates a mix of bidirectional grid long short-term memory (LSTM) and convolutional neural networks (CNNs) to produce realistic synthetic signals. Using various datasets like ECG, EEG, EMG, and PPG, they demonstrated that the synthetic signals closely resembled the original data and outperformed existing models on evaluation metrics such as mean absolute error and root mean square error.
Torfi et al. introduced~\cite{torfi2020corgan} a framework named CorGAN, which employs CNNs to generate realistic synthetic healthcare data while capturing correlations between medical features by merging GANs and autoencoders. In their subsequent study, Torfi et al. proposed a differentially private framework \cite{torfi2022differentially} to generate synthetic data that preserves critical temporal and feature correlations of the original data. They demonstrated superior performance compared to existing models in both supervised and unsupervised settings using various medical datasets.
Fang et al. introduced \cite{fang2022dp} DP-CTGAN, a model combining GANs and differential privacy to address the challenge of limited access to real medical datasets while preserving individual privacy. The proposed model outperformed current leading models under identical privacy constraints. By integrating federated learning, they provided a safer approach to generating synthetic data without transferring locally collected data to a central database. Given the limited availability of annotated PCG data for training, Takezaki et al.~\cite{takezaki2022data} proposed a GAN-based data augmentation method to generate synthetic spectrogram data, which was validated for heart sound detection.

Despite significant progress in synthesizing time series data for healthcare, no prior study has specifically focused on developing models to generate high-quality PCG signals. Furthermore, the effectiveness of existing generative methods in addressing this challenge remains unexplored. This research addresses this gap by systematically evaluating advanced generative models for PCG signal synthesis. Additionally, our work seeks to alleviate the current scarcity of abnormal PCG datasets, which is crucial for advancing diagnostic capabilities in cardiology. We aim to lay the foundation for more effective detection tools for heart murmurs and other cardiac anomalies in future medical research.

\section{Approach}
In this section, we elaborate on the methodologies adopted in our case study. We employ three well-known models in the context of audio signals and time series generation, Wavenet~\cite{oord2016wavenet}, Doppelganger~\cite{lin2019generating}, and DiffWave~\cite{kong2020diffwave}— to train and evaluate their performance in generating high-quality PCG signals. 

\subsection{WaveNet}

WaveNet is a neural network designed for synthesizing highly realistic audio wave forms. It leverages dilated causal convolutions to efficiently capture long-range dependencies in audio data, which is essential for producing coherent speech and music.
Its key innovation lies in the use of causal convolutions to maintain temporal causality, ensuring that future samples do not influence the generation of past samples. The network employs stacked dilated convolutions that exponentially increase the receptive field, allowing the model to capture long-range dependencies in the audio. Gated activation units within the network facilitate selective information propagation, while residual and skip connections prevent the loss of information across layers, thereby addressing the vanishing gradient issue. Finally, a softmax output layer allows WaveNet to produce diverse and novel audio outputs.
The model is employed with one input channel and one output channel, indicating a univariate time series context where the input covariates and the forecast target are singular in nature. The forecasting horizon was set to 14 time steps ahead.
We construct the model with 7 convolutional layers with dilation of size 2 within a single WaveNet block, which determined the model's receptive field. Internally, the model's hidden layers were composed of 89 channels, a specification that determines the breadth of the network's capacity to learn from the data. In parallel, we allocate 199 skip channels. The model is trained for 10 epochs and predicts the last 20\% of the time series data.
\subsection{Doppelganger }
DoppelGANger (DGAN) is a GAN-based model designed to generate synthetic time series data that captures the characteristics of real-world datasets. Technically, it distinguishes itself by leveraging two generators: one for feature generation and another for time series generation, allowing it to model both the static attributes associated with data points and their dynamic temporal behaviors. The architecture incorporates auxiliary features alongside the time series, which enables the network to produce high-dimensional outputs with complex patterns. DoppelGANger employs a specially designed training approach that alternates between the two generators, ensuring that both the feature and time series data are synthesized in harmony. This dual-generator approach, combined with a discriminator that's fine-tuned to differentiate between real and synthetic pairs of features and time series, enables the network to produce synthetic data that's not only diverse but also maintains key statistical properties of the original dataset. The resulting synthetic datasets can then be used for a variety of applications, from data augmentation in machine learning models to privacy-preserving data sharing.
In the experimental configuration of the DoppelGANger model, we established a discriminator network comprising three layers, each with 100 units. The generator network is structured with a single LSTM layer, also consisting of 100 units, which is responsible for capturing the temporal dependencies of the data. Each LSTM cell in the generator is tasked with producing a sequence of 10 time series steps, optimizing the granularity of the generated data. We set the learning rates for both the generator and discriminator networks at 0.0001, ensuring a steady but cautious optimization over the course of training. The model is trained over an extensive run of 10,000 steps, with a substantial batch size of 1,000, to promote model stability and convergence. This setup is chosen to balance the complexity of the model with computational efficiency, allowing for thorough learning without overfitting.
\subsection{DiffWave}
DiffWave, presented in \cite{kong2020diffwave}, leverages the principles of diffusion probabilistic models, and adapts them to the audio domain. The architecture of DiffWave primarily revolves around a two-stage process: a forward process that incrementally adds Gaussian noise to audio signals, and a reverse process where the model learns to denoise these signals, effectively reconstructing the original audio. This denoising process is powered by a deep neural network, possibly featuring convolutional layers, and designed to handle audio data.
The training of DiffWave involves minimizing the difference between the original and reconstructed audio, enabling the model to generate audio that mirrors the complexity and richness of real-world sounds. DiffWave's versatility in synthesizing various audio types, from speech to music, marks it as a notable advancement in generative audio models.

\section{Dataset}
\subsection{Data Description and Analysis}
\label{Data}
We used a publicly available dataset on Physionet Challenges \cite{PhysioNet}. 
The dataset includes one or more heart sound recordings for 942 patients aged 21 years or younger, along with routine demographic information about the patients. Each patient has two labels:
Murmur-related labels indicate whether an expert annotator detected the presence or absence of a murmur in a patient from the recordings or whether the annotator was unsure about the presence or absence of a murmur. Outcome-related labels indicate the normal or abnormal clinical outcome diagnosed by a medical expert \cite{9658215} \& \cite{Reyna2022HeartMD}. In this study, we rely on clinical outcome labels. The PCG signals are sampled at a rate of 4000 Hz and are typically 10 to 30 seconds in duration. Up to five PCG recordings may be collected for each patient from prominent auscultation locations: the pulmonary valve (PV), aortic valve (AV), mitral valve (MV), tricuspid valve (TV), and other locations (Phc). These recordings are captured sequentially using a digital stethoscope.

\begin{figure}[thp!]
    \centering
    \includegraphics[width=0.7\columnwidth]{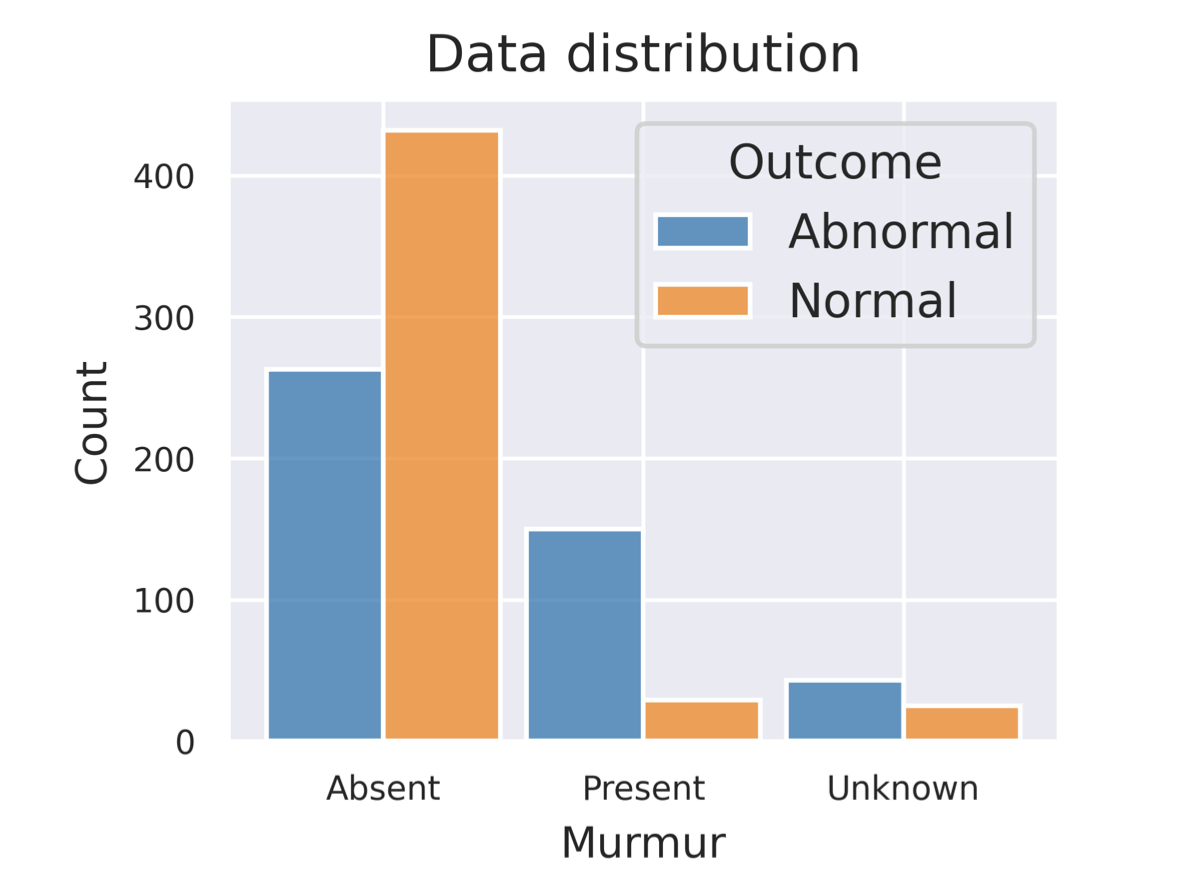}
    \caption{The distribution of the class labels in our dataset.}
    \label{Fig:labels}
\end{figure}

We use a subset of this dataset, focusing on signals from normal subjects, as our goal is to evaluate the performance of existing GAN models in generating normal PCG data. This filtering step resulted in a dataset containing recordings from 486 subjects (see Figure \ref{Fig:labels}).

\subsection{Quality assessments}
In preprocessing PCG signals, we follow a meticulous multi-step approach to ensure data integrity and quality for subsequent analysis. First, we filter the audio recordings of normal subjects based on data description files containing Subject IDs, isolating signals from participants of interest. To remove edge artifacts, the first and last 10\% of the signals are discarded. Next, we apply a rigorous quality assessment protocol adopted from \cite{akram2018analysis}, conducted both with and without the filtering step to evaluate its impact on signal integrity. Finally, we use discrete wavelet transform with Daubechies wavelets at the second level of approximation to extract key features from the PCG signals.

\subsubsection{Root mean square of successive differences (RMSSD)}
We calculate the consecutive differences within the signal and then derive the root mean square of these differences to obtain the RMSSD. A signal is considered appropriate for further processing only if its RMSSD is below a predefined threshold. My experimental analysis indicates that signals with an RMSSD at or above the 0.1 mark are generally indicative of corruption.
\subsubsection{Zero Crossings Ratio}
The number of times the signal intersects the x-axis, referred to as zero crossings, is calculated. This value is normalized by dividing it by the total length of the signal to define the second criterion for signal suitability. Our data analysis reveals that a zero-crossing ratio above 0.3 signifies a noisy signal.

\subsubsection{Ratio of windows having normal number of peaks to total windows in signal}
The signal is segmented into 2200-millisecond windows. Each window is scored `1' if the number of peaks falls within the acceptable range, representing normal peak activity. We calculate the proportion of such windows and apply a threshold to this percentage. For the signal to qualify for further processing in heart sound detection, at least 50\% of its duration must be noise-free. Accordingly, the threshold for this criterion has been set at 50\%.

\subsection{Preprocessing the Data}
The frequency of the main components of PCG signals lies in the 40 to 400 Hz range. To preprocess the data, a first-order Butterworth band-pass filter is used to retain this frequency range. Subsequently, standardization is applied to ensure uniform scaling across all PCG recordings.
Finally, a re-sampling process is applied. The signals are down-sampled at twice the highest frequency to satisfy the Nyquist criterion~\cite{oppenheim1999discrete} and preserve the integrity of the signal's frequency content.

\subsection{Localization and Peak Detection Algorithm}
Leveraging the scipy Python library~\cite{2020SciPy-NMeth}, we pinpoint the peaks associated with the S1 heart sounds. Any S1 peaks that exhibit extreme height values, which could indicate erroneous detection, are discarded to maintain the quality of our dataset.
Following this identification, the signal is normalized to lie within the range [-1, 1]. Afterward, a window is centered around the identified S1 peaks to encompass the S2 peaks as well, standardizing the length of these segments to 110 units.

\section{Metrics and Evaluation Methods}
This section elaborates on the evaluation metrics for our generative approaches.

\subsection{Auto-regressive Metrics} \label{W-eval}
To evaluate the forecasting performance of the Wavenet model, we leverage the following metrics.
\begin{enumerate}

    \item \textbf{Mean Absolute Error(MAE):} is the average of the absolute differences between the predicted and observed true values. The formula for MAE is:
    
    \[\text{MAE} = \frac{1}{n} \sum_{i=1}^{n} \left| y_i - \hat{y}_i \right|\] where n is the number of observations, $y_{i}$ is the true value, $\hat{y}_i$ and is the predicted value.
    \item \textbf{Mean Squared Error (MSE):} is expressed as the average squared difference between the actual observed values and the values predicted by the model. The formula for MSE is as follows.
    
    \[\text{MSE} = \frac{1}{n} \sum_{i=1}^{n} (y_i - \hat{y}_i)^2\] where n is the number of observations, $y_{i}$ is the true value, $\hat{y}_i$ and is the predicted value.
    
    \item \textbf{ 
    Symmetric Mean Absolute Percentage Error (SMAPE):} is an accuracy measure for forecasting models. It addresses some limitations of Mean Absolute Percentage Error by treating over- and under-forecast symmetrically. SMAPE is widely used because it is easy to interpret and is scale-independent. The formula for SMAPE is:

    \[SMAPE = \frac{100\%}{n} \sum_{i=1}^{n} \frac{\left| y_i - \hat{y}_i \right|}{0.5 \times (\left| y_i \right| + \left| \hat{y}_i \right|)}\] 
    where n is the number of observations, $y_{i}$ is the true value,$\hat{y}_i$ and is the predicted value.
    
    \item \textbf{Absolute Coverage Difference (ACD):} is a metric used to evaluate the quality of generated time series data. It measures how well the synthetic data covers the space of the real data. This metric is particularly useful in scenarios where it is important for the generated dataset to accurately represent the diversity and distribution of the original data. In other words, ACD quantifies the difference in how well various aspects or subsets of the real data are captured by the synthetic data.

\end{enumerate}
\subsection{Generative Metrics}
\begin{enumerate}
    \item \textbf{t-Distributed Stochastic Neighbor Embedding (t-SNE):} is a dimensionality reduction technique essential for visualizing high-dimensional datasets in a low-dimensional space such as two or three dimensions. This non-linear approach is particularly adept at revealing intricate structures within data, which linear dimensionality reductions might miss. At its core, t-SNE works by converting distances between data points into probability distributions, and then optimizes these distributions to maintain relative distances in the reduced space, using a Student’s t-distribution to prevent different data clusters from overlapping. Commonly applied in exploratory data analysis, t-SNE helps identify clusters and outliers, facilitating pattern recognition and anomaly detection. However, it is computationally intensive and sensitive to hyperparameter settings like perplexity, which balances local versus global data aspects. Additionally, t-SNE's stochastic nature means visualizations may vary across different runs unless a random seed is set, which is a critical consideration for ensuring reproducible results. There are plenty of studies that used t-SNE as a metric to show the efficacy of their generated models. As a example, we would like to refer to \cite{fahim2023transfusion}.
\item \textbf{Discriminative Score:}
\label{DScore}
This metric is proposed in \cite{yoon2019time}. According to this metric, first, each original sequence is labeled real, and each generated sequence is labeled not real. Then, an off-the-shelf RNN classifier is trained to distinguish between the two classes as a standard supervised task. We then report the classification error on the held-out test set, which is 20\% of the shuffled data. 
This method gives a quantitative assessment of fidelity that samples should be indistinguishable from the real data. Therefore, the model should not be better than a dummy (random) classifier. 

In our experimental framework, we train a binary RNN-based classifier to differentiate between real and synthesized datasets. The classifier's inability to reliably distinguish between the two types of data would be indicative of the high-quality of the synthetic data. The RNN architecture was configured with a single layer comprising 20 hidden nodes, an architecture choice designed to balance model complexity and training efficiency. For the optimization process, we employed the Adam optimizer with a learning rate set to 0.01. The training was conducted over 50 epochs with a batch size of 64. The setup of this experiment was carefully crafted to evaluate the classifier's performance and, by extension, the authenticity of the synthesized data relative to the real data.
\item \textbf{Maximum Mean Discrepancy (MMD):} 
 is a powerful statistical test used to compare the underlying distributions of two different sets of data. Unlike many traditional comparison methods, MMD does not rely on any assumptions about the distributions, making it a non-parametric test. It’s particularly well-suited for scenarios where the data doesn't follow a known distribution, which is common in complex data analysis and machine learning applications.
At the heart of MMD is the concept of embedding distributions into a Reproducing Kernel Hilbert Space (RKHS). Within this space, the mean embedding of each distribution can be directly compared, allowing MMD to quantify the difference between the two distributions as the distance between their respective means in the RKHS.
The MMD metric is calculated as follows:

\[ \text{MMD}^2(X, Y) = \left\lVert \frac{1}{m} \sum_{i=1}^m \phi(x_i) - \frac{1}{n} \sum_{j=1}^n \phi(y_j) \right\rVert^2_{\mathcal{H}} \]

Here, \( X \) and \( Y \) are samples drawn from two distributions, \( m \) and \( n \) are the sample sizes, \( \phi \) represents a feature map to the RKHS, and \( \mathcal{H} \) denotes the RKHS.

In this report, we leverage the MMD metric to assess whether the samples generated by our model differ in a statistically significant way from real data samples. A smaller MMD value suggests that the model-generated data closely approximates the real data in terms of distributional characteristics. This provides a robust methodology for validating the performance of generative models, especially in situations where traditional metrics may fall short. The ability of MMD to work effectively in high-dimensional spaces further underscores its utility in modern data analytic tasks, particularly in the realm of machine learning where understanding and comparing high-dimensional feature distributions is crucial.
We adopted this metric from \cite{esteban2017real}.
\item \textbf{Jensen Shannon Divergence (JSD):} 
is a symmetric and smoothed version of the Kullback-Leibler divergence (KLD). It measures the similarity between two probability distributions and is particularly useful in contexts where the true distribution of the data is unknown or when comparing two different distributions that may have non-overlapping support.

In our analysis, we employed the Jensen-Shannon Divergence as a metric to quantify the dissimilarity between distributions of real and synthesized data. JSD is defined as the average of the KLD from each distribution to the mean of the two distributions. Mathematically, it can be represented as follows:

\[JSD(P \parallel Q) = \frac{1}{2} D(P \parallel M) + \frac{1}{2} D(Q \parallel M)\]

where P and Q are the two distributions being compared, 
M is the point wise mean of P and Q, and D represents the Kullback-Leibler divergence. A notable property of JSD is that it always yields a finite value, even when the distributions do not overlap, due to its inherent smoothing.

The JSD ranges from 0 to 1, where 0 indicates identical distributions, and 1 indicates that the distributions share no support. In our report, we utilize JSD to measure the distance between the real data distribution and the distribution of data generated by our model, offering insight into the performance of our generative processes. A lower JSD value would suggest that the synthetic data closely mirrors the real data, reflecting the efficacy of the generative model in capturing the underlying data distribution. We adopt this metric from previous studies such as \cite{fahim2023transfusion} and \cite{esteban2017real}.
\end{enumerate}
\section{Experimental Results}
\subsection{Wavenet performance}
We evaluate the forecasting performance of the Wavenet model using the metrics discussed in Section \ref{W-eval}. Table \ref{w-met} provides an overview of the model's evaluation. The performance metrics in the aforementioned table suggest that the model closely mimics the characteristics of the original data, generating synthetic data that is highly aligned with the original dataset. For all the metrics mentioned, the less is a better indicator of the prediction's quality.
\begin{table}[thp!]
  \caption{WaveNet's performance metrics in synthesizing PCG signals
  }
  \centering
  \begin{tabular}{@{}ccccc@{}}
    \toprule
    \textbf{Metric}      & \textbf{MAE} & \textbf{MSE} & \textbf{SAMPE} & \textbf{ACD} \\ 
    \midrule
    \textbf{Score Value} & 0.02         & 0.002        & 3.51           & 0.006       \\ 
    \bottomrule
  \end{tabular}
  \label{w-met}
\end{table}

\subsection{Doppelganger GAN performance}
In this section, we report the ability of the Doppelganger GAN model to synthesize high-quality data.

\begin{enumerate}
    \item t-SNE:
    t-SNE, a commonly used method in GAN frameworks and time series data analysis, is applied to reduce temporal N-dimensional sequential data to two dimensions. The result of this conversion is depicted in Figure \ref{Fig:tsne-before}. It is evident that there is significant overlap between the synthetic and real data. Most data points, synthetic (blue) and real (orange), are concentrated in a dense cluster without clear separation.

    \begin{figure}[thp!]
    \centering
    \includegraphics[width=0.9\columnwidth]{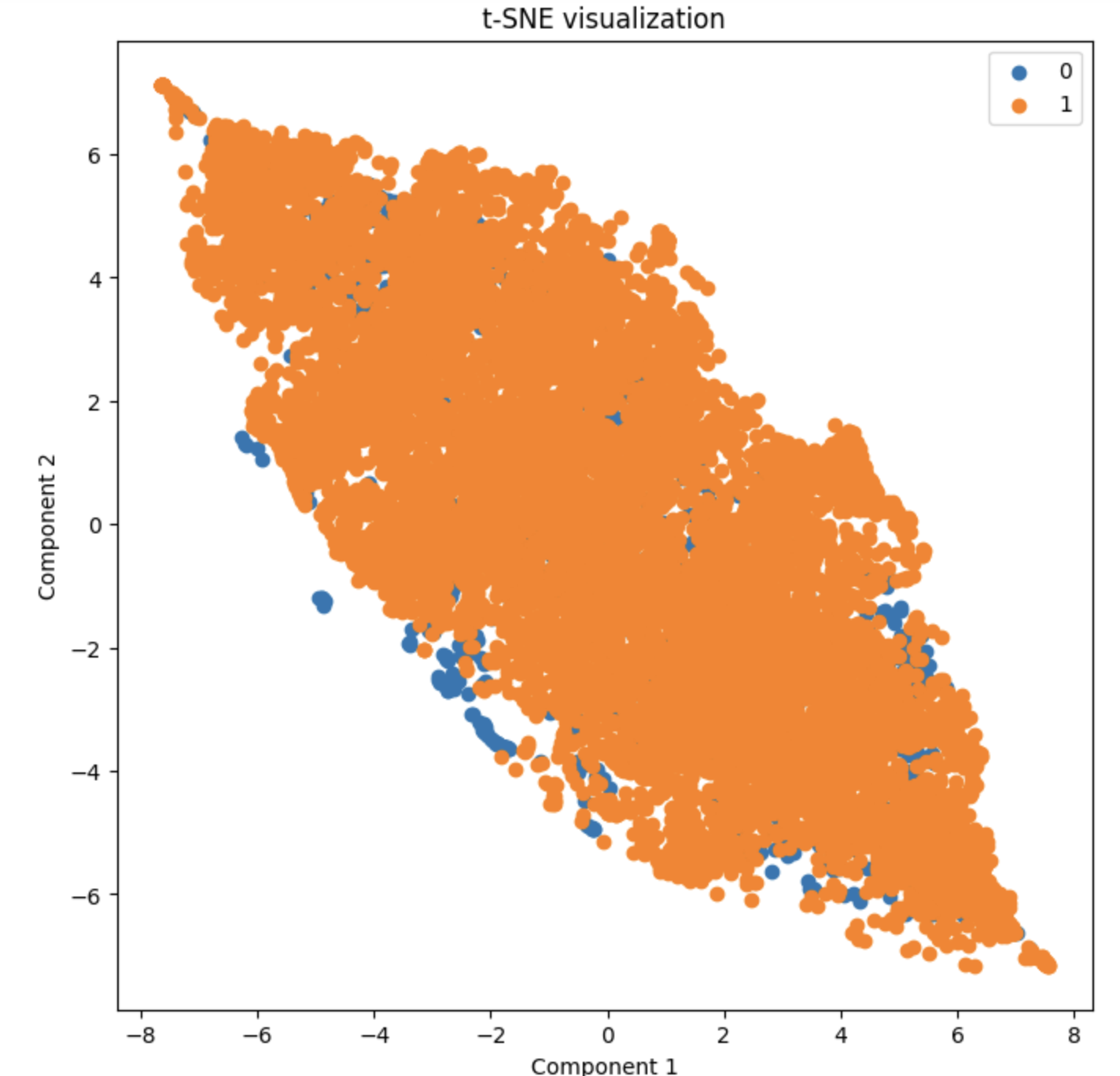}
    \caption{t-SNE analysis on the distributions of real and synthetic PCG sequences generated by DGAN that are displayed in the orange (label 1) and blue dots (label 0), respectively.}
    \label{Fig:tsne-before}
    \end{figure}

    \item Discriminative Score: The RNN binary classifier is trained with 100 epochs and a batch size of 64 on the balanced shuffled dataset of containing the real and synthetic data. The model's performance analysis demonstrates a 52\% accuracy on the validation set, indicating that the classifier is unable to reliably distinguish between real and synthetic data. This finding underscores the high quality of the synthetic data generated by our Doppelganger model, which effectively replicates the characteristics of the authentic dataset.

    \item MMD and JSD Scores:  A JS divergence score close to 0 indicates very similar or nearly identical distributions. A score of 0.014 suggests that the two distributions of the synthetic and the original data are quite similar. MMD  measures the distance between the mean embeddings of two distributions in RKHS. A score of 0.0001  suggests that, in the chosen RKHS, the mean embeddings of the two distributions are almost indistinguishable. According to these two metrics, both low scores indicate that the synthetic data generated by the model is very similar to the real data.
   \begin{table}[thp!]
  \caption{DGAN's Performance metrics in synthesizing PCG signals
  }
  \label{js-mmd}
  \centering
  \begin{tabular}{@{}ccc@{}}
    \toprule
    \textbf{Metric}      & \textbf{JSD} & \textbf{MMD} \\ 
    \midrule
    \textbf{Score } & 0.014        & 0.0001       \\ 
    \bottomrule
  \end{tabular}
\end{table}

\end{enumerate}

\subsection{DiffWave Performance on Synthesizing PCG signals}
In the case of the DiffWave model, we also evaluate its performance in generating healthy PCG signals based on the previously mentioned generative metrics.

\begin{enumerate}
    \item t-SNE:
    The result of converting the real and synthetic data into two dimensions is depicted in Figure \ref{Fig:tsne-diffwave-noraml} for generating PCG sequences. It is easy to observe that 
    there is a significant overlap between the synthetic and real data points. The majority of the data points, both synthetic (blue) and real (orange), are concentrated in a dense cluster without clear separation.
        \begin{figure}
    \centering
    \includegraphics[width=0.9\columnwidth]{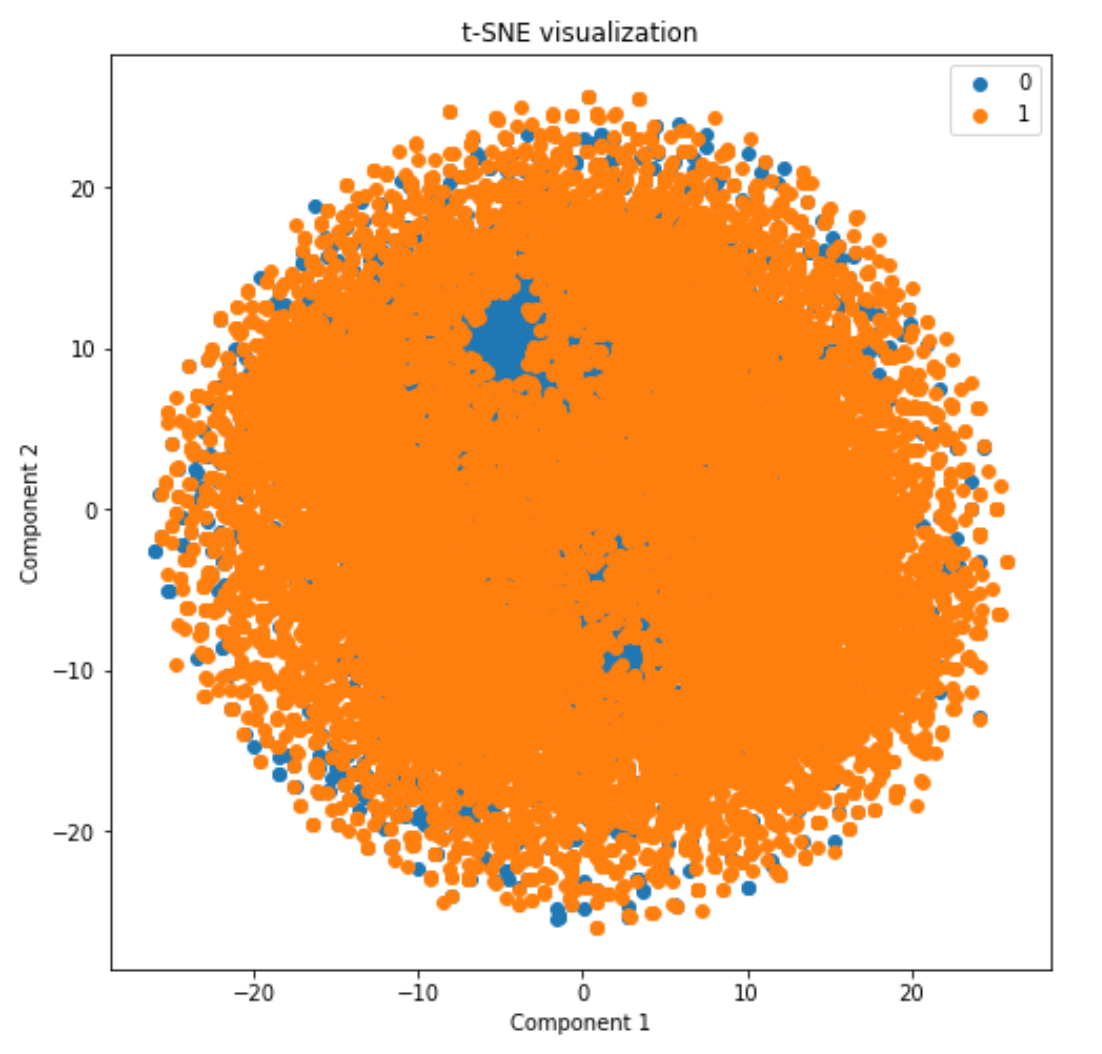}
    \caption{t-SNE analysis on the distributions of real and synthetic Normal PCG sequences generated by DiffWave that are displayed in the orange (label 1) and blue dots (label 0), respectively.}
    \label{Fig:tsne-diffwave-noraml}
    \end{figure}

    \item Discriminative Score:
     We train the RNN classifier described in Section \ref{DScore} with the same hyperparameters and experimental setups. The classifier is trained for 100 epochs with a batch size of 64 on a balanced and shuffled dataset comprising real and synthetic PCG data. The RNN-based classifier achieved 51\% accuracy on the validation set, indicating the model's inability to distinguish between real and synthetic data.

    \item MMD and JSD Scores:
    The MMD and JS scores highlight that the distributions of the real and synthetic data are virtually indistinguishable. The quantitative scores are given in Table~\ref{js-mmd-diffwave}.
    
    \begin{table}[thp!]
  \caption{DiffWave's performance metrics in synthesizing PCG signals. 
  }
  \label{js-mmd-diffwave}
  \centering
  \begin{tabular}{@{}ccc@{}}
    \toprule
    \textbf{Metric}      & \textbf{JSD} & \textbf{MMD} \\ 
    \midrule
    \textbf{Score } & 0.015        & 0.0001       \\ 
    \bottomrule
  \end{tabular}
\end{table}

\end{enumerate}

\section{Conclusion}

In this study, we investigated the feasibility of generating PCG signal data using multiple generative models, specifically focusing on audio and time series data forecasting through WaveNet, Doppelganger GAN, and Diffwave. Our findings demonstrate the feasibility of this approach and our ability to synthesize high-quality healthy PCG data.
To assess the quality of the generated data, we employed a range of widely used metrics, including t-SNE, JSD, MMD, and the discriminative prediction score, all of which are adopted from previous studies in the domain of time series generation and forecasting.
We utilized t-SNE to visually inspect the degree of similarity between the generated data and the original data, providing an intuitive understanding of their similarity. Through MMD and JSD scores, we established that the generated data closely mirrors the characteristics of the original data. Moreover, by using a discriminative prediction pipeline, we demonstrated that a binary classifier struggles to differentiate between real and synthesized data, further emphasizing the high-quality of the generated data.
In our future work, we aim to utilize the proposed method to address the existing data gap for murmured PCG signals to improve the heart murmur detection tools.

\bibliographystyle{IEEEtran}  
\bibliography{Refs}

\end{document}